\documentclass{article}
\usepackage{spconf,amsmath,graphicx}
\usepackage{xcolor}
\usepackage{hyperref} 

\usepackage[numbers]{natbib} 
\usepackage{booktabs}
\usepackage{multirow}
\usepackage{graphicx} 
\usepackage{makecell}
\usepackage{float}
\usepackage{lipsum}
\usepackage{cuted}      
\usepackage{caption}    

\usepackage{makecell,multirow,array}

\newcommand{\HBVCVGGT}{SceneVGGT}

\title{\HBVCVGGT: 
VGGT-based online 3D semantic SLAM
\\for indoor scene understanding and navigation}

\name{
\parbox{\textwidth}{\centering
Anna Gelencsér-Horváth$^{1 2 \star \dagger}$\thanks{$^\star$Corresponding author: \texttt{gha@itk.ppke.hu}}\thanks{$^\dagger$ Equal contribution}
\thanks{Project page \& code: \href{https://hbvc-ai.github.io/SceneVGGT/}{https://hbvc-ai.github.io/SceneVGGT/}}
\qquad 
Gergely Dinya$^{2 \dagger}$ \qquad 
Dorka Boglárka Erős$^{1}$ \qquad 
Péter Halász$^{1}$ \qquad \\
Islam Muhammad Muqsit$^{1}$ \qquad 
Kristóf Karacs$^{1}$ \qquad
}}

\address{$^{1}$ Pázmány Péter Catholic University, Budapest, Hungary \\ 
$^{2}$ Eötvös Loránd University, Budapest, Hungary\\
}

\begin{document}
\maketitle
\begin{abstract}
We present \HBVCVGGT, a spatio-temporal 3D scene understanding framework that combines SLAM with semantic mapping for autonomous and assistive navigation. Built on VGGT, our method scales to long video streams via a sliding-window pipeline. 
We align local submaps using camera-pose transformations, enabling memory- and speed-efficient mapping while preserving geometric consistency. Semantics are lifted from 2D instance masks to 3D objects using the VGGT tracking head, maintaining temporally coherent identities for change detection. As a proof of concept, object locations are projected onto an estimated floor plane for assistive navigation. 
The pipeline's GPU memory usage remains under 17 GB, irrespectively of the length of the input sequence and achieves competitive point-cloud performance on the ScanNet\textit{++} benchmark.
Overall, \HBVCVGGT~ensures robust semantic identification and is fast enough to support interactive assistive navigation with audio feedback.

\end{abstract}

\section{Introduction}
\label{sec:intro}


3D scene understanding is a key enabler for assistive navigation in cluttered, unfamiliar, and constantly changing indoor environments. 
A practical system must build a temporally coherent 3D map that remains stable under viewpoint changes and occlusions, while also capturing semantic information to support meaningful tasks such as identifying free seats or locating relevant objects. 
At the same time, deployment in real-world scenarios requires low memory usage and fast processing, since input arrives as a continuous stream and guidance must be provided real-time.

These requirements are similar to those for an autonomous navigating agent, though there are important differences.
On the one hand, the assistive navigation task is somewhat more relaxed because it is possible to rely on the human operator's intelligence. 
On the other hand, there is a greater uncertainty about the actions that are carried out, which is inherent to human execution and the incorporation of information sensed by the human person. 

Recent feed-forward multi-view transformers such as Visual Geometry Grounded Transformer (VGGT)~\cite{vggt} provide sufficiently accurate depth and camera pose estimates for such a purpose, making them attractive for fast 3D reconstruction. 
However, directly applying VGGT to long sequences is limited by its rapidly increasing memory requirements as the length of the sequence grows as well as the lack of explicit temporal consistency over extended streams. 

\begin{figure}[H]
    \centering
    \includegraphics[width=\linewidth]{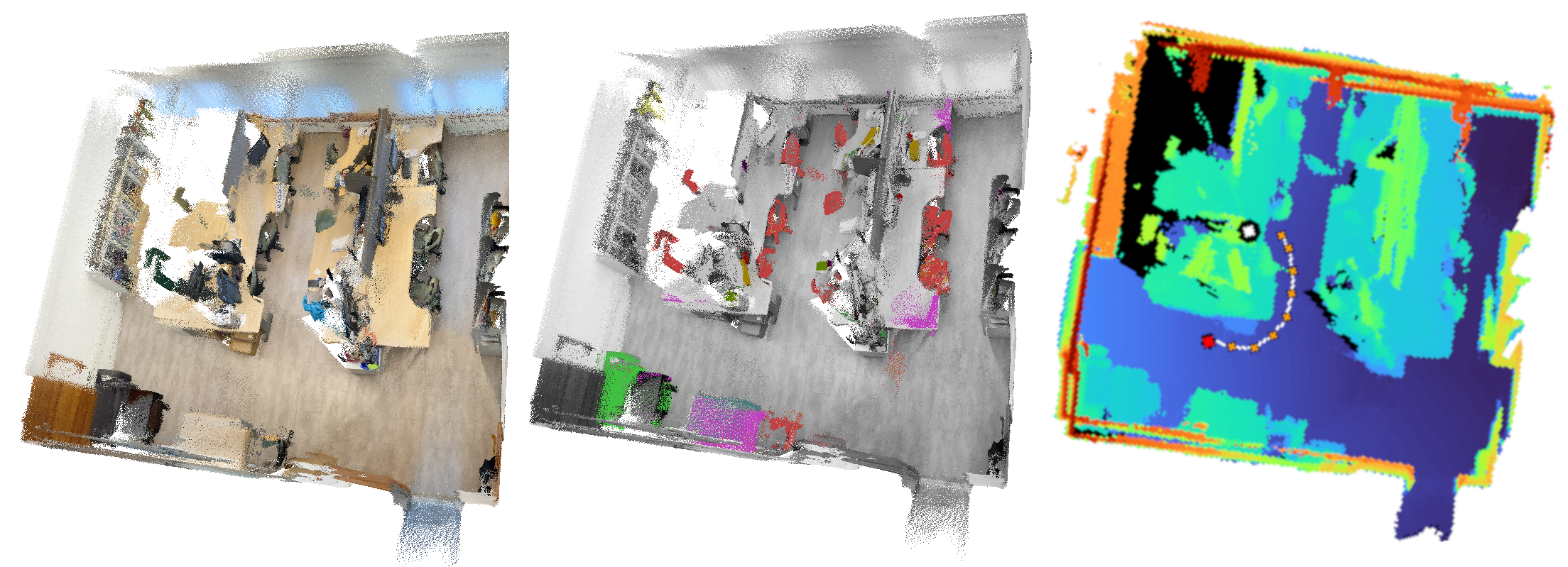}
    \caption{RGB and semantic reconstruction of an in-the-wild scene in an office along with the navigation map and route.}
    \label{fig:placeholder}
\end{figure}

We introduce a memory-efficient streaming semantic SLAM pipeline for indoor sequences, adapting VGGT with a sliding window and camera-pose–based submap alignment, with the following key contributions:
\begin{itemize}
    \setlength{\itemsep}{0pt}
    \setlength{\parskip}{0pt}
    \setlength{\parsep}{0pt}
    \setlength{\topsep}{0pt}
    \item 3D temporally coherent semantic mapping by lifting 2D instance masks into 3D and tracking instances using the VGGT tracking head.
    \item Computationally efficient, temporally consistent change detection enabled by persistent object identities and timestamps.
    \item Floor-plane projection of object locations to support downstream assistive navigation modules.
    \item A proof-of-concept assistive navigation module.
\end{itemize}
An results of the main modules of \HBVCVGGT~ are shown in Fig.~\ref{fig:placeholder} for a sample scene.

\section{Related work}
\label{sec:related}

Feed-forward multi-view transformers made dense 3D reconstruction from a set of RGB images practical without heavy multi-view optimization. 
However, in VGGT, computation and memory usage scale poorly with sequence length, and the method is not directly applicable to sequential data (e.g., streams). 

FastVGGT~\citep{fastvggt} speeds up inference via token merging, while SwiftVGGT~\citep{swiftvggt} proposes a scalable variant for large-scale scenes with faster inference under similar feed-forward assumptions. 
For cross-modal, metric-scale mapping, LiDAR-VGGT~\citep{lidarvggt} fuses LiDAR and RGB in a coarse-to-fine manner to improve global consistency and recover metric scale, at the cost of requiring additional sensing beyond monocular RGB.

To extend dense reconstruction beyond short clips, VGGT-Long~\citep{vggtLong} partitions sequences into overlapping chunks and uses robust alignment plus global adjustment, but remains largely offline. 
VGGT-SLAM and VGGT-SLAM 2.0~\citep{vggtslam, vggtslam2} explicitly formulate dense RGB SLAM by creating local submaps from overlapping windows and optimizing projective SL(4) transforms with loop closure to address uncalibrated monocular ambiguity. 
Related dense/global alignment ideas also appear in learning-based reconstruction systems such as SLAM3R~\citep{slam3r} and multi-view stereo networks with memory mechanisms such as MUSt3R~\citep{must3r}, which aim to scale reconstruction to larger image collections.

For online processing, StreamVGGT~\citep{streamVGGT} replaces global attention with causal attention and maintains a streaming state for efficient inference, however the context can still accumulate over time leading to increased memory overhead and latency.
IncVGGT~\citep{incvggt} makes incremental VGGT more feasible for long sequences via bounded-memory design choices, and InfiniteVGGT~\citep{infinitevggt} targets truly endless streams using a rolling memory with pruning strategies to reduce long-horizon drift.

IGGT~\citep{iggt} learns instance-grounded features, clusters them into 3D-consistent masks, and uses these masks to prompt VLMs/LMMs for open-vocabulary querying and grounding.
However, its multi-frame inference is memory-intensive, exceeding 24GB of VRAM on an NVIDIA RTX 4090 for forward passes of roughly 12--15 frames.

EmbodiedSAM~\citep{embodiedsam} leverages SAM for online 3D instance segmentation, highlighting the promise of foundation models for embodied perception, but its 2D-to-3D lifting remains sensitive to mask layering/temporal consistency.


\section{Methods}
\label{sec:methods}

The proposed \HBVCVGGT~pipeline consists of three main modules, where (1) we apply VGGT with LIDAR grounding to build an incremental 3D map using a sliding-window alignment strategy and pose-graph optimization. 
We (2) integrate semantics into this mapping process by enforcing persistent object identities, ensuring that the same real-world object is assigned a consistent label throughout the entire map. Additionally, if an object is moved or appears, the 3D map is accordingly updated, allowing for a temporally consistent mapping. 
Finally, we (3) demonstrate a proof-of-concept navigation module that exploits a floor-plane projection for dimensionality reduction, enabling efficient assistive or autonomous navigation while leveraging the robustness of the underlying SLAM back-end.

\subsection{Online 3D scene alignment for streams} 
\label{subsec:align}
Operating on a continuous stream without global post-processing, \HBVCVGGT~focuses on pragmatic accuracy over exhaustive reconstruction to serve the requirements of temporally coherent semantic understanding and navigation.
To mitigate the prohibitive compute and memory costs of a naive implementation---which would, for each incoming frame at time $t$, have to re-consider all previous frames $0,\dots,t-1$---we adopt a sliding-window approach. 
The stream is partitioned into contiguous, disjoint blocks of size $n$. Each window contains the current block plus $k\leq n$ frames from the previous block, which act as keyframe anchors for inter-block alignment when running VGGT’s pose and depth heads on the resulting $n+k$ frames.
$n$ and $k$ are chosen to match the available hardware resources.
We have experimented with different values of $k$, and settled with $k=n$.
\footnote{The number of frames of a block is $n=10$ in our setup with an NVIDIA RTX 4090 GPU.} 
Inter-block pose transforms are estimated using the $k$ overlapping anchor frames, enabling us to compose the current block’s relative transforms and align them with the previously accumulated trajectory.
This ensures that consecutive submaps remain aligned to the initial global anchor.
We ground the VGGT-predicted depth estimations with the depth data from the sensor to obtain absolute scale and spatial accuracy, and to enable the system to operate in metric units as required for navigation and egolocation. 
%
%
%
To improve robustness, we construct a combined validity mask for both depth maps that excludes
(i) pixels whose VGGT-predicted depth confidence is below a threshold (we used $c=1.1$) or falls in the lowest $10\%$ of all values 
and
(ii) pixels lying outside the sensor's reliable operating range.
Absolute filtering helps avoiding alignment corruption in the case when most depth estimates are unreliable.
The block scale is obtained as the median of the scale over frames that minimizes the L2 error between the masked predicted depth and the masked LiDAR depth.
This global scale is applied to the translational part of the extrinsic matrix and to the predicted depth maps to align them with the LiDAR metric scale.
To generate the incrementally built point cloud data (PCD), we omit the VGGT PCD head entirely and instead back-project the RGB-D frames using the estimated poses and depth maps.

\subsection{Lifting 2D semantics with change awareness}
\label{subsec:semantics}
We propose a method to represent real-world objects as 3D point clouds where points are labeled with instance types and IDs, maintaining these labels consistently throughout the sequence to define global objects over time.

We perform instance segmentation on 2D RGB frames, 
assigning a mask and a class label to each instance, 
then post-process the masks with a slight erosion to reduce noise during inverse projection into 3D. 
We propagate instance masks from frame to frame 
using the VGGT tracking head, 
leveraging its contextual cues even in challenging scenarios with numerous overlapping and occluding objects.
To enhance computational and memory efficiency, we apply uniform grid-based sampling on each mask and propagate the selected points.
The VGGT tracking head is initialized once for each block, starting at the first non-keyframe. 
%

\begin{figure}[H]
    \centering
    \includegraphics[width=\linewidth]{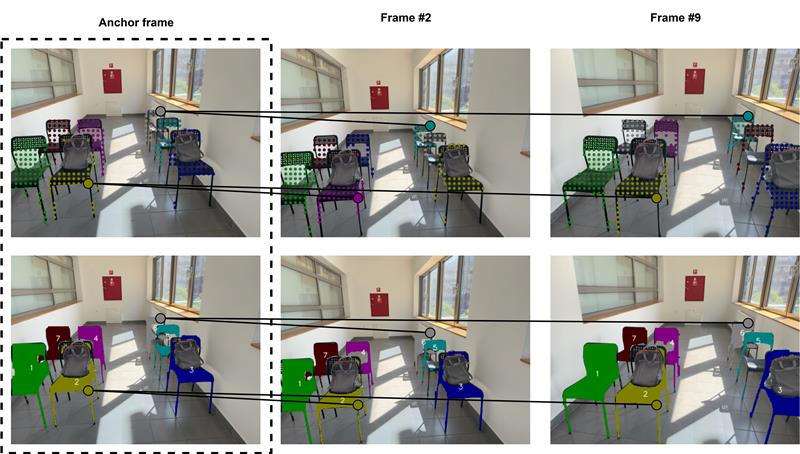}
    \caption{Illustration of the mask sampling and point tracking among frames.}
    \label{fig:instancetrack}
\end{figure}

A tracklet is defined as a set of instance masks that correspond to the same physical object across time. 
New tracklets are initialized only on the first keyframe of each block.
Once a mask is assigned a tracklet ID, all pixels in the mask inherit that ID. 
We uniformly subsample pixels from each labeled mask in the first keyframe and propagate these track-labeled pixels to the remaining frames in the block using the VGGT tracking head. 
In the subsequent frames, we assign each mask the tracklet ID of the majority track label among propagated points falling inside it. Class consistency between the mask and the tracklet is also required. 
Because the keyframes come from the previously processed block, this procedure ensures temporal instance consistency across blocks by construction.
Masks that receive no propagated points from any tracklet (i.e., objects that appear mid-block) are marked as ``untracked'' and 
no new tracklets are initialized for them in the current block.
Tracklets with no assigned observation in a frame are considered inactive. 
Fig.~\ref{fig:instancetrack} illustrates mask sampling and point tracking among frames.

We maintain a persistent memory of previously recognized objects to enable re-identification across temporal gaps and block boundaries. 
Newly initialized tracklets are compared against inactive ones using Chamfer distance by matching the sparse point clouds constructed from the per-frame instance point clouds' median points. 
After computing the average Euclidean distances to the nearest neighbor in the other point cloud in both directions (using KD-trees), we take the minimum of the two directed averages as the final distance. 
This avoids penalizing partial observations.
If the best match is below a threshold (30 cm in our setup) 
the two tracklets are merged. 
Using median points per frame also mitigates projection artifacts occurring in case of 2D masks of objects with holes.

The background and foreground components (instance segmented objects) of the point cloud corresponding to the scene, are maintained separately. 
The 3D position is stored for every point in both point clouds, however, in the foreground cloud the object class label and global instance ID are also included. 
In the 3D cloud we do not store RGB information. 
%
%
%
This design lets us update the 3D representation efficiently by avoiding the costly back-projection of the entire accumulated foreground point cloud upon each update, since instance information is stored directly with the 3D points. 
%

To allow for updating changes in the scene during time 
we exploit that frames arrive sequentially and assign the index within the sequence to each frame.
For each tracklet we maintain the last frame index in which it was observed and a confidence score $c$. 
An object can be in one of three states: (1) \textsc{Recent}, (2) \textsc{Removed}, or (3) \textsc{Retained}. 
Objects observed in the current block are in \textsc{Recent} state with a confidence score of $c=1$. 
At the end of each block, we update all objects by back-projecting their 3D points into the current view to check 
whether they are in the camera’s field of view. 
If the object is in view, we use the depth image to estimate whether it is occluded. 
If it should be observable 
but no detection is associated with it, we apply a linear decay to the confidence value. 
When repeated misses drive $c$ to zero the status of the tracklet is  eventually set to \textsc{Removed}, consistent with a true scene change.  
%
If an object is out of view, it becomes \textsc{Retained} and we maintaint its current $c$.

\subsection{Navigation}
\label{subsec:navigation}

We restrict our setting to indoor environments with a flat floor plane. 
We derive a top-down 2D map by projecting observations onto the floor, reducing navigation to a 2D problem. The floor plane is estimated from the reconstructed 3D point cloud using a RANSAC-based method~\cite{ransac}.

The estimation is carried out after every few blocks to confirm that the currently used reference plane is valid.
Two planes are considered identical if the angle between their normals is below $5^{\circ}$
and the absolute offset difference is below $0.1m$.
If a new estimate does not match the current reference plane, then it is compared to the last $l$ plane estimates. If it matches the majority of them, then we update the reference plane and recompute the navigation maps.
We set $l=5$ for the assistive navigation benchmark, corresponding to five blocks (50 frames, 
$\sim4$ seconds at our frame rate), to reduce the impact of short, abrupt phone movements.

For navigation, we project our 3D point cloud onto a 2D grid.
We continuously update a point density map by dividing the total number of points projected on each grid cell by the number of frames with any observation projected on the cell, yielding a mean map that is less sensitive to object dwell time. 
As long as the reference floor-plane equation remains unchanged, we fuse each block’s point cloud into the existing map using the same origin and scale, producing an incrementally growing map.
If the reference plane changes, all  accumulated points are reprojected onto the new plane.
The map is denoised using a morphological opening operation. 
The map is refreshed in each block to update the changes of object states, ensuring that it remains consistent and reliable for navigation. 

We propose a proof-of-concept navigation module that selects the goal based on semantic information (e.g., finding a seat) by filtering the scene with a semantic mask for the target object class, discarding small connected components to suppress spurious detections, and then choosing the remaining object whose centroid is closest to the user’s current position.
To account for the agent’s non-point-like size, we enlarge occupied cells by morphological dilation to create a spatially extended representation of the obstacles on the map, yielding a conservative free-space map for safe traversal.
Some floor cells may remain unobserved at a given step, since LiDAR observations are incomplete and the map is updated locally in a sliding-window, block-wise manner.
We mark unknown area on the floor as non-traversable in the obstacle map, while handling it separately from detected obstacles by excluding morphological inflation.
Known versus unknown floor is obtained via height-based filtering of the minimum-height map relative to the estimated floor plane, producing a binary known/unknown mask.
When no segmented target is visible, we perform frontier-based exploration~\cite{yamauchi1997frontier, topiwala2018frontier} 
using proximity to the user as a secondary preference, while relying on a human-in-the-loop for local navigation in unknown areas (e.g., using a white cane) to reduce collision risk in the assistive setting.

\section{Results}
\label{sec:results}

\subsection{Pose and reconstruction}

We evaluate pose and reconstruction metrics on the 7-Scenes dataset~\citep{sevenscenes}. 
Following the evaluation protocol of StreamVGGT~\cite{streamVGGT} we evaluated pose accuracy, with the Root Means Square Error~(RMSE) of the Average Trajectory Error~(ATE) 
and the reconstruction, based on Accuracy~(Acc), Completeness~(Compl.) and Normal Consistency~(NC).
We built on the evaluation codes provided in the repository of Infinite VGGT~\cite{infinitevggt}. 
We report average and median values in Table~\ref{tab:vggt_summary_new} compared to existing approaches.
Although, no direct comparison for performance is possible due to hardware differences, we report GPU memory and running speed indicatively.
For InfiniteVGGT, IncVGGT, and SceneVGGT, the reported peak GPU memory is effectively independent of sequence length (i.e., it remains constant with increasing sequence length).

\newcommand{\OOM}{\textcolor{gray}{OOM}}
\newcommand{\na}{\textcolor{gray}{\emph{n/a}}}

\begin{table*}[t]
\centering
\scriptsize
\setlength{\tabcolsep}{3.0pt}
\renewcommand{\arraystretch}{1.18}

\resizebox{\textwidth}{!}{%
\begin{tabular}{|l|c|cc|cc|cc||c|c|c|c|}
\hline
\textbf{} &
\multirow{3}{*}{\makecell{\textbf{ATE} \\ \textbf{(m)}}} &
\multicolumn{6}{c||}{\textbf{Reconstruction metrics for max. 500 frames}} &
\multicolumn{2}{c|}{\makecell{\textbf{GPU}}} & 
\multicolumn{2}{c|}{\makecell{\textbf{Performance for 300 frames}}}  \\
\cline{3-12}
& &
\multicolumn{2}{c|}{\textbf{Accuracy}} &
\multicolumn{2}{c|}{\textbf{Completeness}} &
\multicolumn{2}{c||}{\textbf{NC}} & 
\multirow{2}{*}{\makecell{\textbf{NVIDIA}\\ \textbf{model}}} & \multirow{2}{*}{\textbf{VRAM}} & 
\multirow{2}{*}{\makecell{\textbf{Peak} \\ \textbf{VRAM usage} }} &
\multirow{2}{*}{\makecell{\textbf{Speed } }} \\
\cline{3-8}
& &
\textbf{Mean} & \textbf{Med.} &
\textbf{Mean} & \textbf{Med.} &
\textbf{Mean} & \textbf{Med.} & 
& & & \\
\hline
Infinite VGGT~\cite{infinitevggt} & n/a & 0.043  & 0.018  & 0.025  & 0.005  & 0.561  & 0.593  & A100 & \na           & 14.49 GB               & 5.95 fps \\
\hline
VGGT*         & \na & 0.0055 & \na    & 0.0067 & \na    & 0.948*  & \na    & A100 & 80 GB            & 60+ GB                  & -- \\
\hline
VGGT**       & \na & 0.087  & 0.039  & 0.091  & 0.039  & 0.787  & 0.890  & H100 & 80 GB             & --                      & 8.53 fps \\
\hline
VGGT SLAM*    & 0.067 & 0.031  & 0.0173 & 0.0387 & 0.0173 & 0.612  & 0.675  & RTX 4090& 24 GB      & 24$>$ GB                  & 6.9 fps \\
\hline
StreamVGGT    & \na & \na    & 0.0241 & \na    & 0.0203 & \na    & 0.915  & A800& 40 GB           & 80 GB+                & 2.16--2.2 fps \\
\hline
IncVGGT       & \na & \na    & 0.0266 & \na    & 0.0203 & \na    & 0.901  & A100& 80 GB            & 9 GB                  & 11.74 fps\\
\hline
SceneVGGT     & 0.0628 & 0.0213 & 0.0120 & 0.0446 & 0.0086 & 0.5775 & 0.6186 & RTX 4090 & 24 GB     & 15 GB                   & 7.23 fps \\
\hline
\end{tabular}%
}

\caption{Evaluation of pose accuracy (RMS of the average trajectory error (ATE)) and reconstruction quality (mean distance) of the alignment on the 7-Scenes dataset, as defined in~\cite{streamVGGT} with a stride of 2 and a maximum length of 500 frames. 
For \HBVCVGGT~(alignment only) and VGGT SLAM (SL(4), w=16) own measurements are shown; 
results for IncVGGT, StreamVGGT, and VGGT** (with sparse sampling) are taken from~\cite{incvggt}, 
whereas Infinite VGGT and VGGT* (with a stride of 2) are from~\cite{infinitevggt}. 
For VGGT* in the Normal Consistency (NC) column $NC_1$ values (as defined in~\cite{infinitevggt} are reported).
For runtime and peak GPU memory (VRAM),  measurements are standardized to 300-frame sequences to reduce the impact of hardware memory-capacity differences.}
\label{tab:vggt_summary_new}
\end{table*}


\vspace{-0.4cm}
\subsection{Semantics}
For evaluation we use ScanNet\textit{++}~\cite{scannetpp} scenes from the validation and test splits.
To ensure that the evaluation is independent of a foundation model’s recall on the benchmark, we use the scenes’ ground-truth semantic masks (GT). 
Since the memory footprint of VGGT scales with the number of tracked 3D points per instance, we evaluate under a closed-vocabulary setting and restrict the GT labels to a small set of nine frequent, task-relevant categories that overlap with those of MS COCO dataset~\citep{mscoco}. 
This choice mirrors human attention, which prioritizes a few key targets over clutter, and it keeps inference feasible under bounded memory.
For each 2D frame, we load the ground-truth instance masks for the aforementioned selected nine classes whenever they are present in the image, in an arbitrary order, as if it was a model's prediction.
We measure ID consistency by the dominance ratio: for each ground-truth ID, we take the most frequent non-zero predicted ID across frames as dominant and report its share over all frames of that ground-truth track (consistency without counting objects appearing mid-block).
Additionally, we report a ID consistency with counting the objects where it is already detected semantically mid-block, but has not yet been assigned a tracklet (i.e., predicted ID=0).

We randomly selected office and apartment scenes containing chairs to also provide a proof-of-concept qualitative evaluation for a typical assistive-navigation use case: finding an empty seat.
We provide a visualization video for one scene (Scene \textit{09c1414f1b}) in the supplementary material.

A direct comparison is not available, nevertheless, we report IGGT’s Temporal Success Rate for context. 
IGGT’s scores 98.9\% for TSR, but multi-frame inference is restricted to short clips ($\sim$12 frames within 20 GB of VRAM) and becomes memory-prohibitive beyond that. 
For reference, SpaTracker+SAM and SAM2 achieve 23.68\% and 57.89\% TSR, respectively.
The complete SceneVGGT pipeline however, including instance segmentation, labeling, tracking, and navigation modules add an additional 2 GB of VRAM, bringing the total peak memory usage to 17 GB, independently of the sequence length.

\begin{table}[ht]
\centering
\small
\setlength{\tabcolsep}{8pt}
\begin{tabular}{|l|c|c|c|}
\hline
\textbf{Scene ID} & \textbf{09c1414f1b} & \textbf{25f3b7a318} & \textbf{acd95847c5} \\
\hline
\makecell[l]{Length (no.\\of frames)} & 1073 & 1054 & 1029 \\
\hline
\makecell[tl]{ID cons. w\textbackslash o}  &  90.14\%  &  95.72\%   &  95.56\%\\
\hline
\makecell[tl]{ID cons. w}  &  74.29\%  &  69.29\%   & 65.87\%\\
\hline
\end{tabular}
\caption{Semantic evaluation results on ScanNet\textit{++} with 10 keyframes. {
ID cons. stands for ID consistency, with or without counting objects appearing mid-block.}
Column headers indicate the scene IDs. We evaluate the scenes with a stride of 1.}
\label{tab:semantics_scannet_temporal}
\end{table}

\subsection{Qualitative results}
We also include qualitative visualizations, since numerical scores alone may fail to capture important visual artifacts. 
For 2D segmentation in-the-wild, we use YOLOv9e~\citep{wang2024yolov9} from Ultralytics~\citep{ultralytics_yolo_2023}, pretrained on MS COCO~\cite{mscoco}, and report qualitative results on our own real-world sequences. 
This evaluation is included to illustrate change detection and map updates under natural conditions, which type of events are not covered by the ScanNet++ benchmark.

We report qualitative results for plane estimation and navigation, in the absence of a widely used benchmark for assistive navigation in our setting. 
We apply a scenario of finding an empty seat. 
with the goal is defined as to find the nearest available seat. Since the dataset does not include actions executed in response to the navigation output, the selected goal may switch dynamically between candidates over time. 
Nevertheless, this setup is sufficient to demonstrate the capabilities of the proposed approach.

Runtime is not benchmarked on a standardized public protocol, instead, we provide an indicative measurement on a representative in-the-wild sequence of $1\,500$ frames with a stride 2. 
The alignment stage takes 126.59 s (5.92 fps for the 750 frames), adding semantics ($+$31.35 s) and instance tracking with object and change management ($+$50.79s) yields 210.14s total (3.57 fps), which is sufficient for interactive assistive navigation and leaves headroom for future audio feedback.
We provide illustrative videos as Supplementary material.

\subsection{Limitations and future plans}
While our change-aware object memory supports appearance/disappearance and relocation updates, the current pipeline does not explicitly model continuous object motion or fully dynamic scenes. 
Moreover, we plan to extend change detection beyond recognized instances to also capture updates in unsegmented regions.

\section{Conclusion}
\label{sec:conclusion}
We present \HBVCVGGT, a VGGT-based online 3D semantic SLAM framework that scales to long indoor video streams via a sliding-window design and camera-pose–based submap alignment. 
By integrating LiDAR-derived scale at each step, we effectively mitigate the long-term drift that typically affects monocular-based systems, while preserving metric consistency required for navigation. 
The proposed 2D-to-3D semantic lifting with VGGT tracking enforces temporally coherent instance identities and enables change-aware map updates through a persistent object memory. 
Finally, floor-plane projection provides a compact top-down representation that supports downstream assistive navigation, demonstrating the practical utility of the approach under realistic compute and memory constraints.




\bibliographystyle{IEEEbib}
\bibliography{refs}

@String(CVPR= {IEEE Conf. Comput. Vis. Pattern Recog.})

@String(ECCV= {Eur. Conf. Comput. Vis.})

@String(ICLR = {Int. Conf. Learn. Represent.})

@String(CVPR  = {CVPR})

@String(ECCV  = {ECCV})

@String(ICLR  = {ICLR})

@inproceedings{sevenscenes,
  title={Scene coordinate regression forests for camera relocalization in {RGB-D} images},
  author={Shotton, Jamie and Glocker, Ben and Zach, Christopher and Izadi, Shahram and Criminisi, Antonio and Fitzgibbon, Andrew},
  booktitle={Proceedings of the IEEE Conference on Computer Vision and Pattern Recognition (CVPR)},
  pages={2930--2937},
  year={2013},
  organization={IEEE}
}

@inproceedings{vggt,
  title={{VGGT}: Visual Geometry Grounded Transformer},
  author={Wang, Jianyuan and Chen, Minghao and Karaev, Nikita and Vedaldi, Andrea and Rupprecht, Christian and Novotny, David},
  booktitle={Proceedings of the IEEE/CVF Conference on Computer Vision and Pattern Recognition},
  year={2025}
}

@article{vggtlong,
  title        = {{VGGT-Long}: Chunk it, Loop it, Align it – Pushing {VGGT}'s Limits on Kilometer-scale Long RGB Sequences},
  author       = {Kai Deng and Zexin Ti and Jiawei Xu and Jian Yang and Jin Xie},
  journal      = {arXiv preprint arXiv:2507.16443},
  year         = {2025},
  url          = {https://arxiv.org/abs/2507.16443},
}

@article{vggtslam,
  title        = {{VGGT-SLAM}: Dense {RGB SLAM} Optimized on the {SL(4)} Manifold},
  author       = {Dominic Maggio and Hyungtae Lim and Luca Carlone},
  journal      = {arXiv preprint arXiv:2505.12549},
  year         = {2025},
  url          = {https://arxiv.org/abs/2505.12549},
}

@InProceedings{mscoco,
author="Lin, Tsung-Yi
and Maire, Michael
and Belongie, Serge
and Hays, James
and Perona, Pietro
and Ramanan, Deva
and Doll{\'a}r, Piotr
and Zitnick, C. Lawrence",
editor="Fleet, David
and Pajdla, Tomas
and Schiele, Bernt
and Tuytelaars, Tinne",
title="Microsoft {COCO}: Common Objects in Context",
booktitle="Computer Vision -- ECCV 2014",
year="2014",
publisher="Springer International Publishing",
address="Cham",
pages="740--755",
isbn="978-3-319-10602-1"
}

@misc{ultralytics_yolo_2023,
  title        = {Ultralytics {YOLO}},
  author       = {Jocher, Glenn and Qiu, Jing and Chaurasia, Ayush},
  year         = {2023},
  month        = jan,
  note         = {Version 8.0.0. License: AGPL-3.0},
  url          = {https://ultralytics.com},
  howpublished = {\url{https://github.com/ultralytics/ultralytics}},
  organization = {Ultralytics}
}

@article{wang2024yolov9,
  title={{YOLOv9}: Learning What You Want to Learn Using Programmable Gradient Information},
  author={Wang, Chien-Yao  and Liao, Hong-Yuan Mark},
  journal = {arXiV},
  booktitle={arXiv preprint arXiv:2402.13616},
  year={2024}
}

@misc{fastvggt,
      title={{FastVGGT}: Training-Free Acceleration of Visual Geometry Transformer}, 
      author={You Shen and Zhipeng Zhang and Yansong Qu and Liujuan Cao},
      year={2025},
      eprint={2509.02560},
      archivePrefix={arXiv},
      primaryClass={cs.CV},
      url={https://arxiv.org/abs/2509.02560}, 
}

@article{streamVGGT,
      title={Streaming {4D} Visual Geometry Transformer}, 
      author={Dong Zhuo and Wenzhao Zheng and Jiahe Guo and Yuqi Wu and Jie Zhou and Jiwen Lu},
      journal={arXiv preprint arXiv:2507.11539},
      year={2025}
}

@article{slam3r,
  title   = {{SLAM3R}: Real-Time Dense Scene Reconstruction from Monocular {RGB} Videos},
  author  = {Liu, Yuzheng and Dong, Siyan and Wang, Shuzhe and Yin, Yingda and Yang, Yanchao and Fan, Qingnan and Chen, Baoquan},
  journal = {arXiv preprint arXiv:2412.09401},
  year    = {2024}
}

@article{must3r,
  title   = {{MUSt3R}: Multi-view Network for Stereo {3D} Reconstruction},
  author  = {Cabon, Yohann and Stoffl, Lucas and Antsfeld, Leonid and Csurka, Gabriela and Chidlovskii, Boris and Revaud, Jerome and Leroy, Vincent},
  journal = {arXiv preprint arXiv:2503.01661},
  year    = {2025}
}

@article{embodiedsam,
  title   = {EmbodiedSAM: Online Segment Any 3D Thing in Real Time},
  author  = {Xu, Xiuwei and Chen, Huangxing and Zhao, Linqing and Wang, Ziwei and Zhou, Jie and Lu, Jiwen},
  journal = {arXiv preprint arXiv:2408.11811},
  year    = {2024}
}

@article{swiftvggt,
  title   = {{SwiftVGGT}: A Scalable Visual Geometry Grounded Transformer for Large-Scale Scenes},
  author  = {Lee, Jungho and Lee, Minhyeok and Yang, Sunghun and Kang, Minseok and Lee, Sangyoun},
  journal = {arXiv preprint arXiv:2511.18290},
  year    = {2025}
}

@article{lidarvggt,
  title   = {{LiDAR-VGGT}: Cross-Modal Coarse-to-Fine Fusion for Globally Consistent and Metric-Scale Dense Mapping},
  author  = {Wang, Lijie and Guo, Lianjie and Xu, Ziyi and Wang, Qianhao and Gao, Fei and Chen, Xieyuanli},
  journal = {arXiv preprint arXiv:2511.01186},
  year    = {2025}
}

@article{iggt,
  title   = {{IGGT}: Instance-Grounded Geometry Transformer for Semantic {3D} Reconstruction},
  author  = {Li, Hao and Zou, Zhengyu and Liu, Fangfu and Zhang, Xuanyang and Hong, Fangzhou and Cao, Yukang and Lan, Yushi and Zhang, Manyuan and Yu, Gang and Zhang, Dingwen and Liu, Ziwei},
  journal = {arXiv preprint arXiv:2510.22706},
  year    = {2025}
}

@article{infinitevggt,
  title   = {{InfiniteVGGT}: Visual Geometry Grounded Transformer for Endless Streams},
  author  = {Yuan, Shuai and Yang, Yantai and Yang, Xiaotian and Zhang, Xupeng and Zhao, Zhonghao and Zhang, Lingming and Zhang, Zhipeng},
  journal = {arXiv preprint arXiv:2601.02281},
  year    = {2026}
}

@inproceedings{incvggt,
  title     = {{IncVGGT}: Incremental {VGGT} for Memory-Bounded Long-Range {3D} Reconstruction},
  author    = {Anonymous},
  booktitle = {ICLR 2026 Conference Submission},
  year      = {2026},
  url       = {https://openreview.net/forum?id=CezA1eLa1Y}
}

@article{ransac,
  author  = {Fischler, Martin A. and Bolles, Robert C.},
  title   = {Random Sample Consensus: A Paradigm for Model Fitting with Applications to Image Analysis and Automated Cartography},
  journal = {Communications of the ACM},
  volume  = {24},
  number  = {6},
  pages   = {381--395},
  year    = {1981},
  month   = jun,
  doi     = {10.1145/358669.358692}
}

@misc{vggtslam2,
      title={{VGGT-SLAM 2.0}: Real time Dense Feed-forward Scene Reconstruction}, 
      author={Dominic Maggio and Luca Carlone},
      year={2026},
      eprint={2601.19887},
      archivePrefix={arXiv},
      primaryClass={cs.CV},
      url={https://arxiv.org/abs/2601.19887}, 
}

@inproceedings{scannetpp,
  title={Scannet++: A high-fidelity dataset of {3D} indoor scenes},
  author={Yeshwanth, Chandan and Liu, Yueh-Cheng and Nie{\ss}ner, Matthias and Dai, Angela},
  booktitle={Proceedings of the IEEE/CVF International Conference on Computer Vision},
  pages={12--22},
  year={2023}
}

@article{topiwala2018frontier,
  title={Frontier based exploration for autonomous robot},
  author={Topiwala, Anirudh and Inani, Pranav and Kathpal, Abhishek},
  journal={arXiv preprint arXiv:1806.03581},
  year={2018}
}

@inproceedings{yamauchi1997frontier,
  author    = {Brian Yamauchi},
  title     = {A frontier-based approach for autonomous exploration},
  booktitle = {Proceedings of the 1997 IEEE International Symposium on Computational Intelligence in Robotics and Automation (CIRA'97), Monterey, CA, USA},
  year      = {1997},
  pages     = {146--151},
  doi       = {10.1109/CIRA.1997.613851},
}

\end{document}